\documentclass{egpubl}
\usepackage{eg2024}
 
\ConferencePaper        
\usepackage[T1]{fontenc}
\usepackage{dfadobe} 

\BibtexOrBiblatex
\usepackage[backend=biber,bibstyle=EG,citestyle=alphabetic,backref=true]{biblatex}
\electronicVersion
\PrintedOrElectronic
\ifpdf \usepackage[pdftex]{graphicx} \pdfcompresslevel=9
\else \usepackage[dvips]{graphicx} \fi

\usepackage{egweblnk} 
\usepackage{booktabs}
\usepackage{lineno}
\usepackage{amsfonts}
\usepackage{amsmath}
\usepackage{enumitem}
\usepackage{multirow}
\usepackage{array}
\usepackage{arydshln}
\usepackage{xcolor}
\usepackage{caption}
\usepackage{tabularx}
\newcommand{\blue}[1]{\textcolor{blue}{#1}}

\title {DivaTrack: Diverse Bodies and Motions from Acceleration-Enhanced Three-Point Trackers}

\author[Yang et al.]
{\parbox
{\textwidth}
{\centering Dongseok Yang$^{1}$, Jiho Kang$^{1}$, Lingni Ma$^{2}$, Joseph Greer$^{2}$, Yuting Ye$^{2}$ and Sung-Hee Lee$^{1}$}
\\
{\parbox
{\textwidth}
{\centering $^{1}$Korea Advanced Institue of Science and Technology, $^{2}$Meta Reality Labs}}}

\begin{document}

\teaser{
 \includegraphics[width=0.85\linewidth]{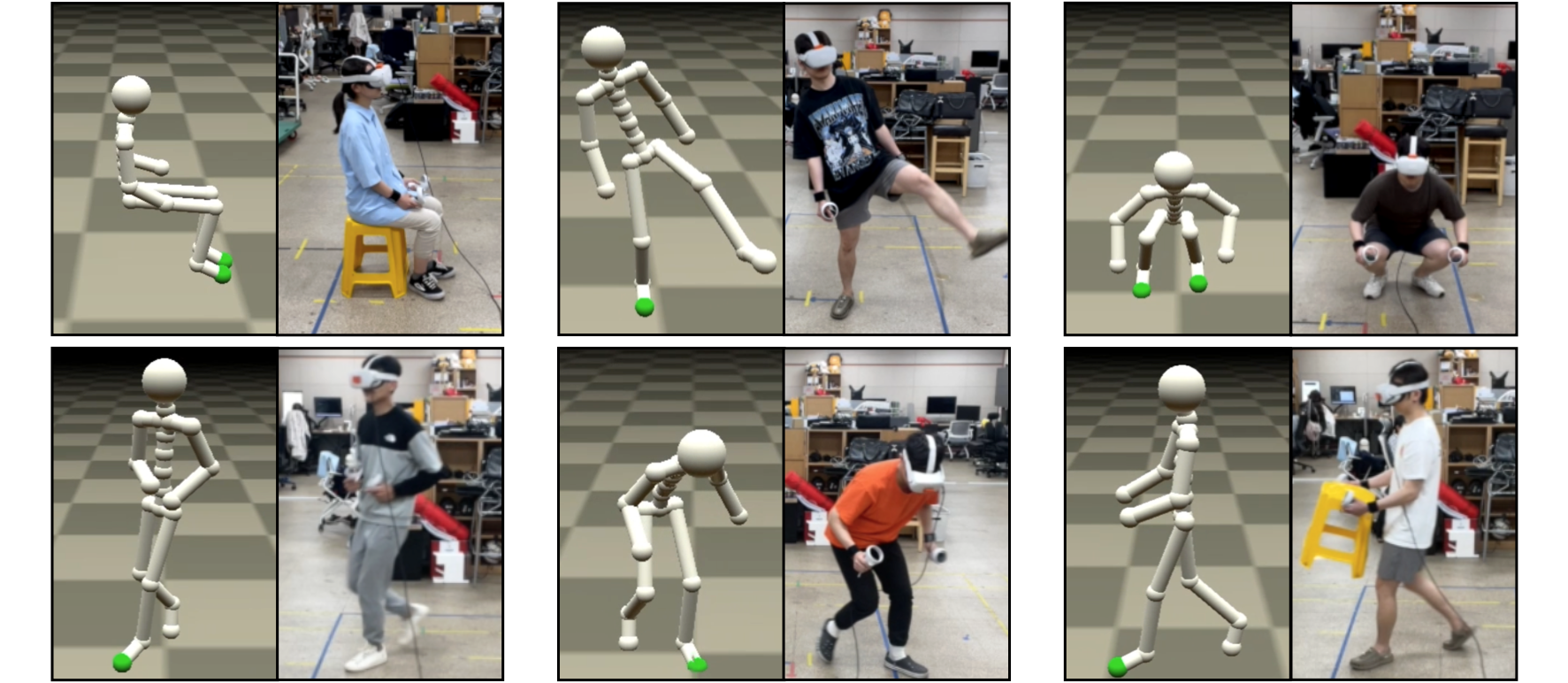}
 \centering
  \caption{Our method generates full-body motions for diverse body proportions and activities from only three-point trackers and IMUs.}
\label{fig:teaser}
}

\maketitle

\begin{abstract}
Full-body avatar presence is crucial for immersive social and environmental interactions in digital reality. However, current devices only provide three six degrees of freedom (DOF) poses from the headset and two controllers (i.e. three-point trackers). Because it is a highly under-constrained problem, inferring full-body pose from these inputs is challenging, especially when supporting the full range of body proportions and use cases represented by the general population. In this paper, we propose a deep learning framework, DivaTrack, which outperforms existing methods when applied to diverse body sizes and activities. We augment the sparse three-point inputs with linear accelerations from Inertial Measurement Units (IMU) to improve foot contact prediction. We then condition the otherwise ambiguous lower-body pose with the predictions of foot contact and upper-body pose in a two-stage model. We further stabilize the inferred full-body pose in a wide range of configurations by learning to blend predictions that are computed in two reference frames, each of which is designed for different types of motions. We demonstrate the effectiveness of our design on a large dataset that captures 22 subjects performing challenging locomotion for three-point tracking, including lunges, hula-hooping, and sitting. As shown in a live demo using the Meta VR headset and Xsens IMUs, our method runs in real-time while accurately tracking a user's motion when they perform a diverse set of movements.

\end{abstract}  

\section{Introduction}
Self-embodied digital humans are a cornerstone of Virtual, Augmented, and Mixed Reality (VR/AR/MR) applications as tracking the users' full-body motion is essential for natural interactions in the virtual environment. However, it is a challenging task with information available from today's VR/AR/MR devices, which typically only provide six degrees of freedom (DOF) pose data of the head and hands, i.e. three-point input. Although some headsets are equipped with cameras, they are often limited in field of view and resolution, unsuited for full-body pose estimation. An ideal solution would rely on on-device sensors alone to produce plausible full-body motions for any user performing any daily activities.

Recent work tackles this challenge with deep learning methods that directly map sparse three-point inputs to full-body motion \cite{jiang22avatarposer, aliakbarian22flag}, or to joint torques for simulating the full-body motion \cite{winkler2022questsim, ye2022neural3points}. They have demonstrated success for simple locomotion, but face degraded quality in less common activities such as running backward and carrying objects, or in cases where the lower-body pose is ambiguous with respect to the three-point input, such as kicking, lunging, crouching, or simply looking around while walking. Additionally, they only target an average body size and therefore cannot handle large variations in body proportions.

In this paper, we propose a novel model, DivaTrack, for generating full-body motions from three-point tracker inputs and IMU signals. DivaTrack is designed to work in real-time and is capable of handling diverse body shapes and motions. A lightweight calibration process is first employed to determine the user's body proportions from a few dedicated poses. Our algorithm then infers the full-body poses that apply to the estimated body proportion. We incorporate several insights into the design of our system, so it can better generalize to a wide variety of body sizes and activities.

First, we augment the three-point six DOF input with linear accelerations from Inertial Measurement Units (IMUs) attached to the headset and wrists as input to our system. IMUs are ubiquitous in consumer electronic devices \cite{Mollyn2023IMUPoser} and are already available in all VR devices. They are currently primarily used to infer orientation, but we find that their high-frequency linear acceleration signals have a strong correlation with impacts on the body, such as foot contact events. In fact, body-mounted IMUs are widely used for gait event detection in the field of biomechanics \cite{moe1998new, lee2010use, giandolini2016foot, chew2018estimating, patoz2022single, day2021low}. Foot contacts have been shown to improve the generation of plausible lower body motions \cite{yang21lobstr, tseng22edge}, so we predict them from the input with linear accelerations and use them for full-body pose prediction.

Second, we find that a crucial ingredient to generalization is the choice of reference coordinate frame for data normalization \cite{Ma:2019:TRD}. A common choice in locomotion research is to choose the root coordinate frame of the character and set its origin on the ground plane. This is not directly applicable to three-point tracking because the root motion is unknown. A reasonable alternative is to use the headset coordinate frame. We observe that it performs well when the user's head is aligned with their direction of movement, but causes issues with motions such as walking backward and looking around while walking. On the other hand, a movement direction aligned with a past trajectory excels in these cases but causes problems if the subject is stationary. Therefore, we devised a two-stream architecture that learns to blend results from these two complementary coordinate frames based on the input.

Finally, we split the motion network into an upper-body model (UC-Model) and a lower-body model (L-Model). The UC-Model is a regression model because its output is relatively well-constrained from three-point and IMU inputs. The L-Model is a Conditional Variational Autoencoder (CVAE)~\cite{sohn2015cvae} conditioned on the upper-body pose and contact states predicted from the UC-Model. The CVAE models the ambiguity and correlation between its input and the lower-body. This design allows us to swap out the upper-body pose module, e.g. with a simple inverse kinematics (IK) solver, without changing the lower-body generation.

To verify the above design choices, we collected a large motion dataset with synchronized ground-truth body motions and IMU signals using the Xsens system~\cite{xsens08}, from a diverse pool of subjects performing challenging everyday activities. Even though there already exist body motion datasets that come with IMU data, such as TotalCapture \cite{trumble17totalcapture} and HPS \cite{guzov2021human}, these are not designed to showcase use cases and challenges of three-point tracking and have limited subject diversity. In our dataset, we specifically design movement protocols to demonstrate difficult scenarios using three-point inputs, such as when the inputs have weak correlations to the lower body, or when common assumptions break down (e.g. lying down and rolling on the floor). Testing on the full range of our subject base and difficult motions such as lunges and hula-hooping confirms the effectiveness of our solution. We further demonstrate the applicability of our methods to realistic scenarios by developing a real-time live demo with a Meta VR headset and Xsens IMUs.

To summarize, this paper has the following main contributions:

\begin{itemize}
\item Enhancing three-point tracker signals with IMU accelerations for more accurate foot contact state and full-body pose generation. 
\item A combination of upper-body inference and conditional lower-body generation models to match the given three-point constraints and produce plausible full-body motion.
\item A novel two-stream blending architecture to utilize complementary reference coordinates for a diverse spectrum of motions.
\item A lightweight calibration process to accommodate diverse body proportions.
\item A large human motion dataset ($16.5$ hours) that includes synchronized ground truth body motion with IMU signals, 22 different body proportions, and diverse motion categories.
\end{itemize}
\section{Related Work}
Human motion tracking is widely studied in computer graphics and vision communities. Fully vision-based methods use monocular or multiple cameras to track allocentric motions of single or multiple people~\cite{goel2023humans, ye2023slahmr, khirodkar2023egohumans, cai2023smplerx, diogo23eg, rong2021frankmocap, shimada20physcap, kanazawa2019learning}. 
With the increasing popularity of AR/VR, understanding self-body poses from wearable sensors has received renewed attention. Recent work explores cameras mounted on the head~\cite{wang2023scene, hakada2022unrealego, tome2020selfpose, luo21dynamics}, wrist~\cite{bodytrack2022} and both~\cite{ahuja22controllerpose}. In this paper, we are interested in sparse body-worn trackers and the remainder of this section focuses on prior work that leverages wearable sensors. In addition, we cover literature related to generative motion synthesis, which the methods in this work draw heavily from.

\subsection{Inertial-only tracking}
Fully inertial-based human motion tracking has been in use for multiple decades. Commercial products, e.g., XSens~\cite{xsens08} and Rokoko are widely tested systems, capable of high-quality motion captures in the wild. However, these solutions are highly encumbering, requiring users to wear 17 IMUs on their bodies. Several methods have been proposed to reduce this number. The pioneering work by Marcard~et~al.~\cite{marcard17sip} recovers full-body pose from 6 IMUs via batch optimization over the entire sequence. As the method is not suitable for online applications, recent learning-based methods focused on statistical human motion modeling. Huang~et~al.~\cite{huang18dip} uses a bidirectional recurrent neural network (BiRNN~\cite{schuster97birnn}) with LSTMs~\cite{hochreiter97lstm} to achieve coherent prediction. A similar approach is investigated by Nagaraj~et~al.~\cite{nagaraj20rnn}. Both methods focus on local poses without solving for the global position of the wearer. TransPose~\cite{yi21transpose} addresses this issue by regressing the body velocity from foot contact and IMU data and fusing the outputs using contact predictions. These works are extended in PIP~\cite{yi22pip} to reduce artifacts by refining the motion with physical constraints, similar to~\cite{shimada20physcap}. EgoLocate~\cite{egolocate23tog} further improves global tracking accuracy by fusing with vision-based tracking from a head-mounted camera. Inspired by the success of transformer networks~\cite{vaswani17transformer}, TIP~\cite{jiang2022tip} proposes a simple attention-based network with autoregression that eliminates delay with improved accuracy.

\subsection{Tracking with sparse six DOF input.}
The prevalence of consumer devices in the AV/VR industry has increased the accessibility of head-mounted devices (HMDs) and controllers that provide six DOF pose estimates. This introduces a novel problem setup: to infer full-body pose from the given poses of the head and hands, so-called three-point tracking. Early attempts to solve this problem include CoolMoves~\cite{ahuja21coolmoves}, which
uses hand velocity and acceleration to retrieve and blend the top-K body poses on a limited activity set. Ponton~et~al.~\cite{ponton22mmvr} extend this work by experimenting with a diverse animation database. LoBSTr~\cite{yang21lobstr} adds the pelvis pose to develop a hybrid solution, where the upper-body is solved with IK and the lower-body and foot contacts are predicted via gated recurrent units (GRUs)~\cite{chung2014gru}. Following this work, Ye~et~al.~\cite{ye2022neural3points} uses two GRUs to predict the upper and lower body independently and adopt reinforcement learning (RL) to train a pose correction policy from physical constraints. Similarly, QuestSim~\cite{winkler2022questsim} and QuestEnvSim~\cite{Lee2023Questenv} learn a policy that generates joint torques to drive a physics simulation of body motion that matches the inputs. AvatarPoser~\cite{jiang22avatarposer} show transformer networks are efficient in learning the full-body motion, which is further improved by EgoPoser~\cite{jiang2023egoposer} with a SlowFast~\cite{feichtenhofer19slowfast} module and body shape prediction. Dittadi~et~al.~\cite{dittadi21iccv} show that a variational auto-encoder (VAE)~\cite{kingma14vae} trained with full-body pose can generate
plausible motion from incomplete three-point input. Similarly, Milef~et~al.~\cite{milef23eg} shows a conditional VAE can produce diverse poses from four-point input. 

\subsection{Neural generative model for motion synthesis.}
Human motion synthesis has an extensive body of prior work. Early research explored probabilistic approaches, e.g., principal component analysis (PCA)~\cite{safonova04tog, chai05siggraph,
liu06pca}, Gaussian mixture models (GMMs)~\cite{min12gmm} and Gaussian
processes~\cite{grochow04, wang08gp, levine12gp}. Pioneered by Holden et al.'s work on motion manifolds~\cite{holden2015learning}, recent
work leverages modern deep generative neural architectures.
Generative adversarial networks (GANs~\cite{goodfellow14gan}) have been
adopted by multiple methods, e.g., in~\cite{ferstl19gesture}
to condition body gestures on speech, in~\cite{wang19gans} to synthesize and
control body motions, in~\cite{starke20gan} to generate interactive movements and in~\cite{li22ganimator} to synthesize motions from a sample sequence. VAE is another widely used model that enables probabilistic sampling to generate motion output. For example, DeepPhase~\cite{starke22deepphase} leverages the frequency domain to learn periodic motion from unstructured data. Moreover, CVAEs~\cite{sohn2015cvae} have been shown to be powerful architectures for motion synthesis, where the condition can be motion history~\cite{ling20character}, audio~\cite{lee19neurips, li20vae}, action types~\cite{petrovich21cvae} and three-point input~\cite{milef23eg}. As alternatives, normalizing flow and neural distance fields are adopted by MoGlow~\cite{henter20moglow} and Pose-NDF~\cite{tiwari2022pose}, respectively. Following the recent success of diffusion models for visual synthesis, Tevet et al.~\cite{tevet22mdm, shafir2023human}, Zhang et al.~\cite{zhang22motiondiffuse}, EDGE~\cite{tseng22edge}, MoFusion~\cite{dabral2022mofusion} and EgoEgo~\cite{li2023ego} show diffusion-based motion synthesis driven by text, music or head motion. The recent work AGRoL~\cite{du2023avatars} conditioned a diffusion model with three-point input and showed remarkable offline motion generation.

\section{D\MakeLowercase{iva}T\MakeLowercase{rack}}

\begin{figure*}[ht]
  \centering
  \includegraphics[width=\linewidth]{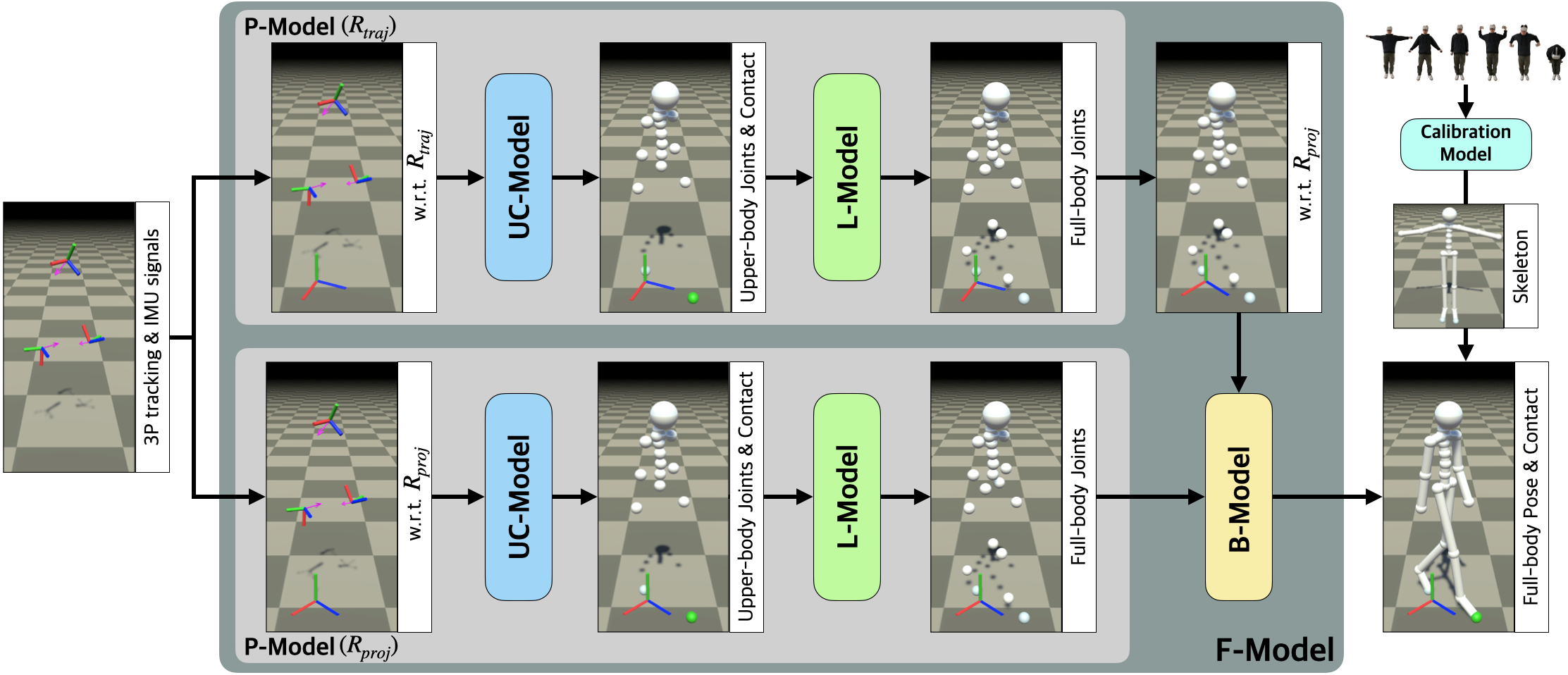}
  \caption{Model overview. Prior to tracking, a learned calibration model estimates the bone offsets from calibration poses once per user. Given three-point six DOF poses and IMU measurements, a full-body tracking network predicts the per-frame motion. This network consists of two streams of identical design to transform inputs into different coordinates. The UC-model predicts the upper-body and foot contacts, which are used to condition the L-model to generate the lower-body. Finally, a blending network fuses the predictions, combined with calibrated offsets to output the final full-body motion.}
  \label{fig:overview}
\end{figure*}

Figure \ref{fig:overview} summarizes our framework to estimate full-body motion from the six DOF poses and linear accelerations of three-point trackers. At the beginning of a session, we use a one-time calibration module to infer the user's body skeleton from six poses (Section~\ref{subsec:BoneLengtH}). We then estimate the user's full-body pose at every frame using the F-Model (Full-body Model) that contains two parallel streams of identically-structured P-Models (Pose Model). A P-Model predicts a full-body pose in two stages: the upper-body {\em inference} and contact prediction (Section \ref{subsec:UpperbodyContact}), followed by the lower-body {\em generation} (Section \ref{subsec:Lowerbody}). Each P-Model is trained with a different reference frame suitable in complementary configurations (Section \ref{subsec:RefereneFrame}). The two P-model outputs are fused together via a motion blending network (B-model). The joint rotations from the blended pose are applied to the calibrated skeleton to output the final full-body pose (Section \ref{subsec:MotionBlending}).

\subsection{Network Input and Output}
\label{subsec:data}
The input state from a tracker $\textbf{x = \{p,\ q,\ v,\ a\}} \in \mathbb{R}^{21}$ is composed of position $\textbf{p} \in \mathbb{R}^3$, rotation $\textbf{q} \in \mathbb{R}^6$, velocity $\textbf{v} \in \mathbb{R}^9$, and linear acceleration $\textbf{a} \in \mathbb{R}^3$. Rotation $\textbf{q}$ follows the 6D representation commonly used in deep learning \cite{zhou2019rotation}. Velocity $\textbf{v}$ is approximated by finite difference of \blue{$(\textbf{p}_t,\ \textbf{q}_{t})$ and $(\textbf{p}_{t-1},\ \textbf{q}_{t-1})$}. We denote the three tracker states as ${\textbf{x}^H,\ \textbf{x}^L,\ \textbf{x}^R}$, for the head, left hand, and right hand, respectively. Our models take as input a trajectory of tracker states within a time window of length $l+1$ ($l=28$ in this work). The end effector input window at time $t$ is $\textbf{X}^{EE}_t = \{\textbf{x}^H_{t-l},\ \textbf{x}^L_{t-l},\ \textbf{x}^R_{t-l},\ \ldots,\ \textbf{x}^H_{t},\ \textbf{x}^L_{t},\ \textbf{x}^R_{t}\} \in \mathbb{R}^{(l+1)\times63}$.

We use a template skeleton of 23 joints, out of which 15 joints belong to the upper-body including the root joint (pelvis), and 8 joints belong to the lower-body. The full-body output $\textbf{y}_{full} = \{ \textbf{y}_{upper},\ \textbf{y}_{lower} \} \in \mathbb{R}^{207}$ is composed of the transformation $\{\textbf{p,\ q}\}$ per joint, with $\textbf{y}_{upper} \in \mathbb{R}^{15\times9}$ and $\textbf{y}_{lower} \in \mathbb{R}^{8\times9}$. In addition, we output the contact probabilities of both feet $\textbf{y}_{contact} \in \mathbb{R}^{2}$. We denote a trajectory of output with $\textbf{Y}_{*,\ t} = \{\textbf{y}_{*,\ t-l},\ \ldots,\ \textbf{y}_{*,\ t}\}$, and the corresponding ground truth with ${\hat{\textbf{Y}}_{*,\ t}}$. We may omit the time index for clarity when the context is clear.

\subsection{Reference Coordinate Frame}
\label{subsec:RefereneFrame}
\begin{figure}
  \centering
  \includegraphics[width=0.9\linewidth]{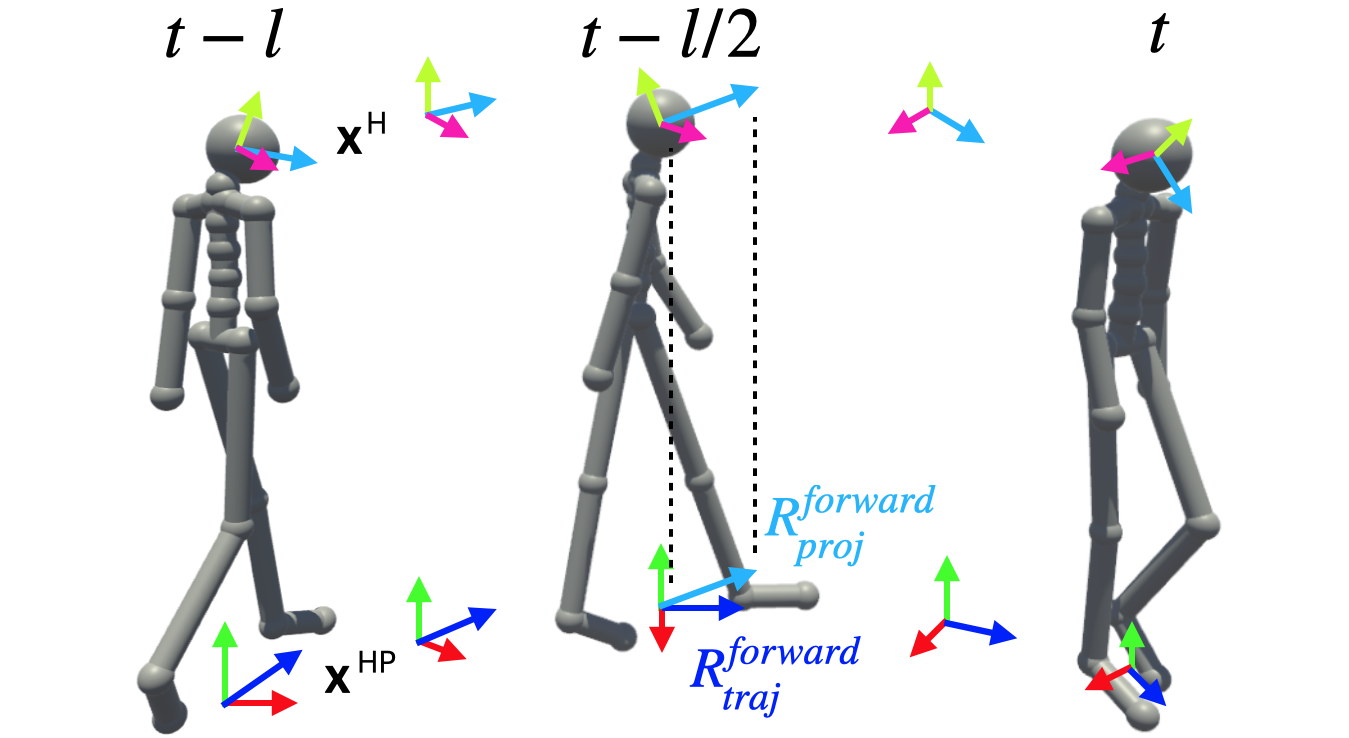}
  \caption{Reference frame definitions. Given a window of head poses $\mathbf{x}_{H}$ from $[t-l,\ t]$, we project the head pose to the ground to obtain $\mathbf{x}_{HP}$ per time-step. $\mathbf{x}_{HP}$ at the middle frame $t-\frac{l}{2}$ is used as the reference, $R_{proj}$, whereas $R_{traj}$ is rotated about the vertical axis to align with their trajectory heading.}
  \label{fig:refdef}
\end{figure}

We use two different coordinate frames, $R_{proj}$ and $R_{traj}$, to represent network input and output. $\textbf{x}^{HP}$ is the world coordinates of the head joint $\textbf{x}^{H}$ projected to the ground plane. It is similar to the commonly used ground-projected root coordinate \cite{starke2019neural}. However, because we don't have access to the ground truth root joint at inference time, we approximate it with the head transformation. The head projection frame $R_{proj}$ equals $\textbf{x}^{HP}$ at time $t-\frac{l}{2}$. Head trajectory frame $R_{traj}$ has the same position as $R_{proj}$ but differs in its forward direction. Specifically, as illustrated in Figure \ref{fig:refdef}, the forward direction of $R_{traj}$ is the normalized vector pointing from the position of $\textbf{x}^{HP}_{t-l}$ to $\textbf{x}^{HP}_{t}$.

\subsection{Skeleton Calibration}
\label{subsec:BoneLengtH}
\begin{figure}
\centering
\includegraphics[width=0.95\linewidth]{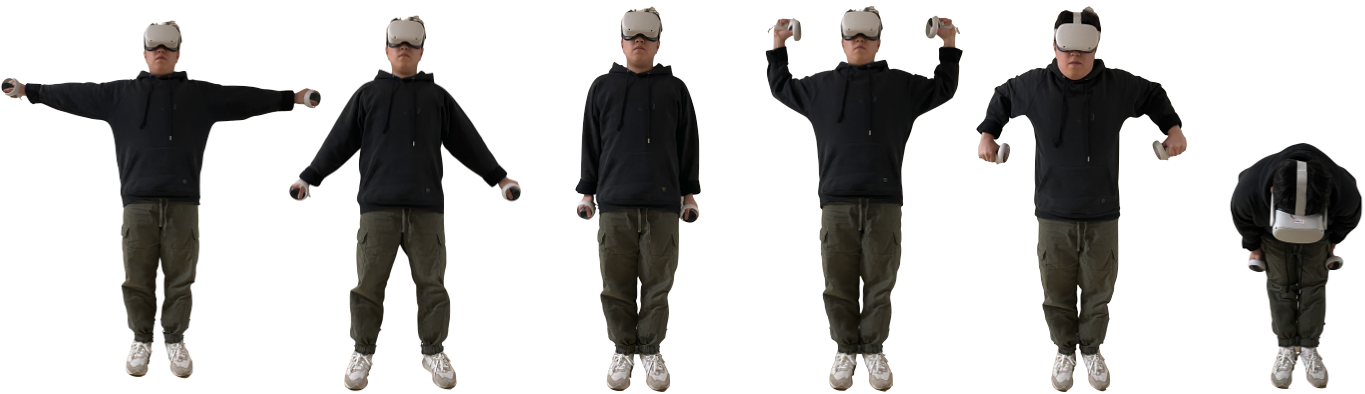}
\caption{Poses for skeleton calibration. From the left, T-pose, A-pose, I-pose, Elbows-bent-up, Elbows-bent-down, and 90$^{\circ}$ bow.}
\label{fig:Cal}
\end{figure}

To calibrate a user's body proportions, we trained a simple 2-layer MLP model to predict the joint positions in our template skeleton model from six calibration poses (Figure \ref{fig:Cal}). The input to this model concatenates the position and orientation of the three trackers for all six poses as $\textbf{x}_{cal} \in \mathbb{R}^{6\times3\times9}$. The output consists of 3D offsets of 22 joints (excluding the pelvis), $\textbf{y}_{cal} \in \mathbb{R}^{22\times3}$. Each input pose is represented in its respective $\textbf{x}^{HP}$. The model is trained to minimize the L2 distance between the predicted and the ground truth offsets, $\mathcal{L}_{cal} =  $$ \Vert \textbf{y}_{cal} - \hat{\textbf{y}}_{cal} \Vert _2$. At run time, a user performs the six poses once and the calibration model estimates the corresponding skeleton dimensions for the final pose reconstruction.

\begin{figure*}
\centering
\includegraphics[width=0.85\linewidth]{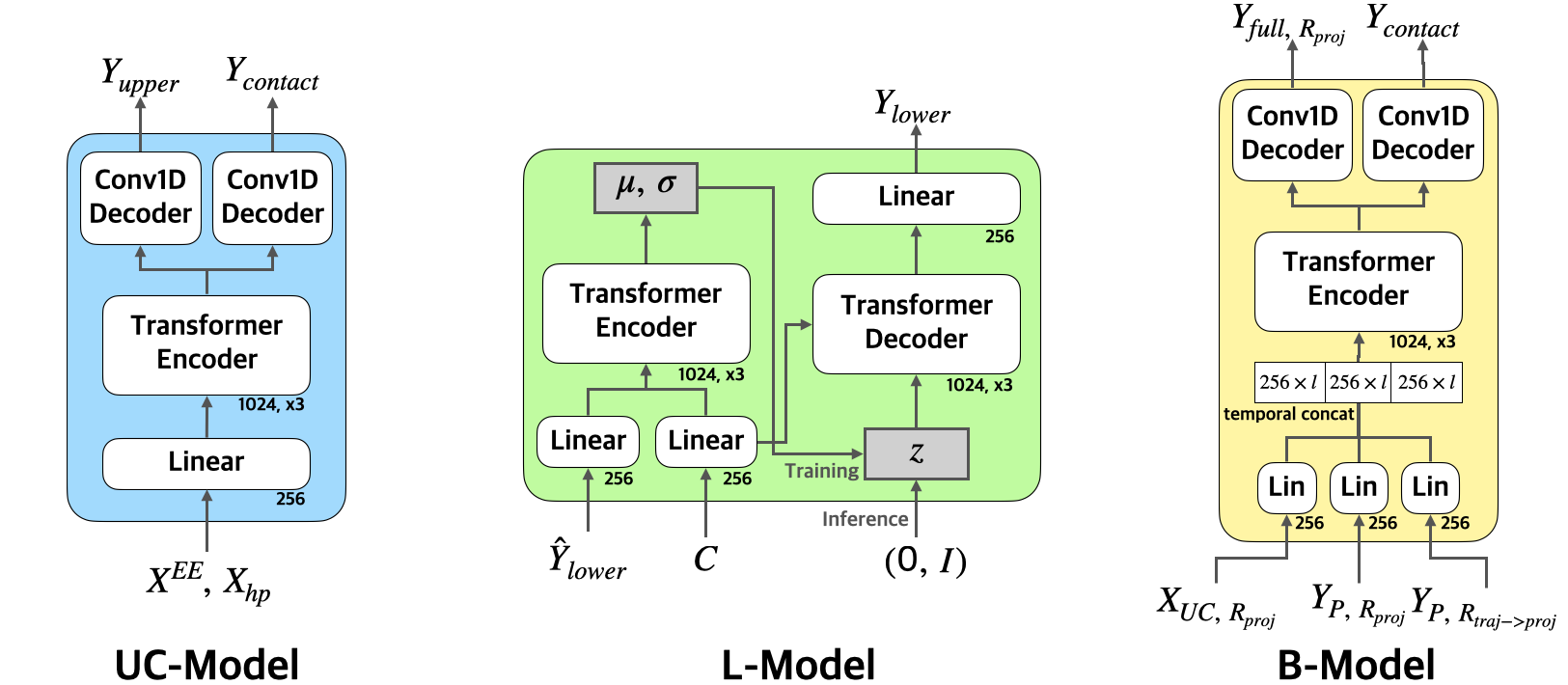}
  \caption{Network architectures. The UC-model is a transformer encoder followed by parallel convolutional decoders that predict upper-body joint rotations and foot contacts. The L-Model is a CVAE, where a transformer encoder learns the lower body latent space conditioned on the upper body prediction and lower body history. The B-model also uses a transformer encoder on learned linear embeddings of the UC-model input and two stream P-model outputs, $Y_{P} = \{Y_{upper},\ Y_{lower},\ Y_{contact}\}$, with convolutional decoders to produce the final full-body and foot contact output.}
  \label{fig:uclb}
\end{figure*}

\subsection{Upper-body and Contact Prediction} \label{subsec:UpperbodyContact}

The Upper-body and Contact Prediction Model (UC-Model) takes $\{\textbf{X}^{EE},\ \textbf{X}^{HP}\} \in \mathbb{R}^{(l+1)\times(63+18)}$ as input, and outputs $\{\textbf{Y}_{upper},\ \textbf{Y}_{contact}\} \in \mathbb{R}^{(l+1)\times(135+2)}$. $\textbf{X}^{HP} = \{\textbf{x}^{HP}_{t-l},\ \ldots, \textbf{x}^{HP}_{t}\}$ is a trajectory where $\textbf{x}^{HP}$ contains the position, orientation, and velocity of the head joint projected to the ground plane ($y=0$). It is derived from $\textbf{x}^H$, but we empirically found that providing this extra data made the network predictions less likely to violate ground constraints. The UC-Model consists of a linear embedding layer, a Transformer encoder \cite{vaswani17transformer}, and two convolutional decoders to estimate an output trajectory of joint transforms and foot contact probabilities, respectively.
It is trained to jointly minimize the L1 error of predicted upper-body joint transformations $\mathcal{L}_{upper} = \Vert \textbf{Y}_{upper} - \hat{\textbf{Y}}_{upper} \Vert _1$, the L2 error of joint velocities $\mathcal{L}_{\Delta upper} = \Vert \textbf{Y}_{\Delta upper} - \hat{\textbf{Y}}_{\Delta upper} \Vert _2$, and the Binary Cross Entropy between predicted and ground truth foot contact probabilities $\mathcal{L}_{contact} = BCE(\textbf{Y}_{contact},\ \hat{\textbf{Y}}_{contact})$.

\subsection{Lower-body Generation} \label{subsec:Lowerbody}
The Lower-body Generation Model (L-Model) is a CVAE~\cite{sohn2015cvae}, using Transformers in both the encoder and the decoder as in TransformerVAE~\cite{petrovich21cvae}. The encoder takes two streams of information: the lower-body joints $\hat{\textbf{y}}_{lower,\ t} \in \mathbb{R}^{72}$ to be constructed, and a conditional vector $\textbf{C} = \{\textbf{Y}_{upper,\ t},\ \textbf{Y}_{lower,\ t-1},\ \textbf{Y}_{contact,\ t}\} \in \mathbb{R}^{(l+1)\times209}$. They are encoded into parameters of a Gaussian distribution $\mathcal{N}(\mu,\ \sigma)$, from which we can sample a latent vector $\textbf{z} \in \mathbb{R}^{256}$. The decoder then reconstructs $\textbf{y}_{lower,\ t}$ from $\textbf{z}$ and the conditional vector $\textbf{C}$. The model is trained to jointly minimize the Kullback-Leibler divergence between the latent distribution and multivariate Gaussian distribution $\mathcal{L}_{KL} = D_{KL}(\mathcal{N}(\mu,\ \sigma)\Vert\mathcal{N}(\textbf{0},\ \textbf{I}))$, the L1 reconstruction loss of the lower-body pose $\mathcal{L}_{lower} = \Vert \textbf{Y}_{lower} - \hat{\textbf{Y}}_{lower} \Vert _1$, and the L2 error of joint velocities $\mathcal{L}_{\Delta lower} = \Vert \textbf{Y}_{\Delta lower} - \hat{\textbf{Y}}_{\Delta lower} \Vert _2$.

At training time, we apply an aggressive dropout rate of $90\%$ to $\textbf{Y}_{lower,\ t-1}$ to avoid overfitting. In addition, we randomly flip $10\%$ of the ground truth foot contact labels to help the model be more robust to prediction errors from the UC-Model. At inference time, we supply the decoder only with a $\textbf{z}$ sampled from $\mathcal{N}(\textbf{0},\  \textbf{I})$, and the conditional vector $\textbf{C}$ from predictions.

\subsection{Full-body Pose} \label{subsec:PoseModel}
The UC-Model and the L-Model together constitute the full-body Pose Model (P-Model), whose input is tracker state $\textbf{X}^{EE}_t$ and output is a trajectory of full-body poses $\textbf{Y}_{full,\ t}$ and foot contact probabilities $\textbf{Y}_{contact,\ t}$. The output is represented in the same reference coordinates as the input. Instead of using the world coordinate as a reference, the P-Model can transform its input to an input reference frame, such as $R_{proj}$ or $R_{traj}$.

\subsection{Motion Blending} \label{subsec:MotionBlending}
Our Full Model (F-Model) runs two P-Models in parallel, one using $R_{traj}$ as the reference, and the other using $R_{proj}$. Depending on the input signals, one of these two outputs could outperform the other. For example, we observed that $R_{proj}$ excels for static movements with stable reference orientation, and $R_{traj}$ produces more accurate predictions when the whole body moves independently of the head orientation (e.g., side or backward stepping). To take advantage of the better between the two, we use a Motion Blending Model (B-Model) to blend them based on the input signals.

The B-Model takes three streams of input, all transformed to the $R_{proj}$ reference frame so the final output pose is also represented in $R_{proj}$. In addition to the two outputs from P-Models, it also takes the same input signals as the UC-Model $\textbf{X}_{UC} = \{\textbf{X}^{EE},\ \textbf{X}^{HP}\} \in \mathbb{R}^{(l+1)\times(63+18)}$ as guidance for the motion blending process. These three sets of input are first transformed by their respective embedding layers to be the same dimension of $(l+1)\times256$. They are then concatenated along the time axis to form the input tensor for the Transformer encoder. Lastly, two convolutional decoders, similar to those of the UC-Model, transform the encoded feature into the final trajectory $\{\textbf{Y}_{full},\ \textbf{Y}_{contact}\}$. The B-Model is trained to minimize the L1 distance between predicted and ground truth full-body pose $\mathcal{L}_{full} = \Vert \textbf{Y}_{full} - \hat{\textbf{Y}}_{full} \Vert _1$, the L2 error of joint velocities $\mathcal{L}_{\Delta full} = \Vert \textbf{Y}_{\Delta full} - \hat{\textbf{Y}}_{\Delta full} \Vert _2$, and Binary Cross Entropy between blended and ground truth contact probabilities $\mathcal{L}_{contact,\ B} = BCE(\textbf{Y}_{contact},\ \hat{\textbf{Y}}_{contact})$. In addition, we further encourage consistency between final foot contact probabilities and predicted foot locations with a contact consistency loss $\mathcal{L}_{consist} = \Vert (FK(\textbf{y}_{full}^{i}) \\ - FK(\textbf{y}_{full}^{i-1})) \cdot \textbf{y}_{contact}^{i} \Vert_2$. as in EDGE \cite{tseng22edge}.

The result of all processing described thus far is a trajectory of transformations for all 23 joints of the skeleton. To produce a final pose, we first compute the pelvis position by running FK from the input head transformation along predicted spinal joint transformations, using calibrated joint offsets. Then, the final full-body pose is reconstructed from this computed pelvis position, output joint rotations from $\textbf{y}_{full,\ t}$, and the calibrated skeleton.

\subsection{Training}
We train all components of the F-Model end-to-end from scratch with a combined loss function as follows:
\begin{equation}
\begin{aligned}
\mathcal{L}\ = &\ \lambda_{pose} (\mathcal{L}_{upper,\ R_{proj}} + \mathcal{L}_{lower,\ R_{proj}} \\ &
\ \ \ \ \ \ \ \ + \mathcal{L}_{upper,\ R_{traj}} + \mathcal{L}_{lower,\ R_{traj}} 
+ \mathcal{L}_{full})
\\ &\ + \lambda_{vel}(\mathcal{L}_{\Delta upper,\ R_{proj}} + \mathcal{L}_{\Delta lower,\ R_{proj}} \\ &
\ \ \ \ \ \ \ \ + \mathcal{L}_{\Delta upper,\ R_{traj}} + \mathcal{L}_{\Delta lower,\ R_{traj}} 
+ \mathcal{L}_{\Delta full})
\\ &\ + \lambda_{contact}(\mathcal{L}_{contact,\ R_{proj}} 
+ \mathcal{L}_{contact,\ R_{traj}} 
+ \mathcal{L}_{contact,\ B})
\\ &\ + \lambda_{consist}(\mathcal{L}_{consist})
\\ &\ + \lambda_{KL}(\mathcal{L}_{KL,\ R_{proj}} 
+ \mathcal{L}_{KL,\ R_{traj}}),
\end{aligned}
\end{equation}

where $\lambda_{pose},\ \lambda_{vel},\ \lambda_{contact},\ \lambda_{consist},$ and $\lambda_{KL}$ are $1,\ 0.1,\ 1\times10^{-3},\ 1\times10^{-5},$ and $ 1\times10^{-3}$, respectively. We minimized the L1 norm in pose losses to avoid averaging similar outputs in a diverse dataset. For velocity and contact consistency loss, we minimized L2 errors for better overall accuracy. We find the velocity losses effective in reducing jitter in the motion, and the contact consistency loss helps to reduce foot sliding. We used Adam optimizer with a learning rate of $1\times10^{-4}$ and batch size of $256$. Training is run on one Nvidia RTX 4090 GPU for about $16.5$ hours with $60,000$ iterations.

\subsection{Dataset}
We captured our dataset using the Xsens Awinda system \cite{xsens08} of 17 IMU sensors attached to the subjects' bodies. The captured IMU data is processed by the Xsens Analyze Pro software to produce ground truth body motions and foot contact labels. We generate synthetic tracker signals by computing the transformations on the head joint and the wrist joints of both hands from the body motion. This practice is consistent with existing work \cite{jiang22avatarposer, du2023avatars} without access to data from commercial VR devices, as the transformations from three-point trackers are often clean and noise-free \cite{winkler2022questsim}.

To improve data consistency between training and testing, we deliberately put IMU sensors at the three tracker locations at capture time, so the acceleration signals from these sensors are similar to those in the live demo. It is of critical importance that we use linear accelerations from the sensors for training because they capture high-frequency signals that correspond to impacts from contact events. In fact, the Xsens software relies on sensor accelerations to annotate ground truth contact labels. Such information is lost in the synthetic acceleration derived from finite differences of poses.

\begin{table}[h]
\begin{tabularx}{0.45\textwidth}{|X|} 
\hline
\textbf{Range of Motion} \\
\hline
T-pose, A-pose, Idle, Elbows bent up \& down, Bow \\
Stretch arms, Look, Roll head, Touch toes, Hands on waist \\
Twist torso, Hula hoop, Lean upper-body \\
Lunge, Squat, Jumping Jack, Kick, Lift knee \\
Turn in place, Walk in place, Run in place \\
Conversational gestures \\
\hline
\textbf{Locomotion} \\
\hline
Normal walk, Walk with free upper-body motions \\
Normal Jog, Jog with free upper-body motions \\
Normal Run, Run with free upper-body motions \\
Normal Crouch, Crouch with free upper-body motions \\
Transitions with changing pace \\
Moving backward with changing pace \\
Jump, Running Jump \\
\hline
\textbf{Object Interaction} \\
\hline
Sit on a chair / couch / sofa / stool / bed \\
Free upper-body motions while sitting \\
Lie down on a sofa / bed, Work at an office desk \\
Moving boxes, Open the door/windows \\
Turn on/off lights, Watch TV, Wash dishes, Cook \\
\hline
\end{tabularx}
\caption{Motion categories in DivaTrack dataset. In addition to standard locomotion patterns with arm swinging, our dataset includes diverse upper-body actions (i.e. drinking water, airplane arms, and clapping) during locomotion. These varied upper-body motions pose challenges to the three-point tracking problem.}
\label{tab:datacomposition}
\end{table}

Our dataset consists of $22$ subjects (height${}_{cm}\ h \sim \mathcal{N}(173.77,\ 8.98^{2})$,\ $159 \leq h \leq 192$) performing around 35 diverse actions, including in-place body exercises, locomotion at different speeds and styles, and interaction with objects and furniture. Table \ref{tab:datacomposition} presents details on motion categories in the dataset. The data was captured in two environments: an empty mocap stage and a furnished apartment. The mocap stage enables a broad range of dynamic motions based on specific protocols, while the apartment focuses on interactions in a natural indoor environment. In total, the dataset contains $772$ motion clips and $16.5$ hours of data. We hold out four subjects for testing and use $18$ subjects for training and validation. The test subjects' body shape (height${}_{cm}$: $[167,\ 171.5,\ 178,\ 192]$) and motions are completely unseen to the trained models in Section \ref{sec:evaluation}.

\section{Evaluation}
\label{sec:evaluation}
We evaluate our method both qualitatively and quantitatively to support the following claims:

\begin{itemize}

\item Our method robustly synthesizes full-body motion from sparse three-point signals in real-time, for individuals with diverse body proportions.

\item Our predict-and-generate approach supports a wider range of motion activities compared to state-of-the-art, including static motion, basic locomotion, and object interactions.

\item Our reference system blending approach improves the estimation quality for the diversity of motions we support.

\item Using foot contact probabilities as conditional input improves the quality of lower-body pose generation.

\item Incorporating IMU linear accelerations in the input leads to more accurate foot contact predictions, which in turn improves the quality of synthesized motions.

\end{itemize}
 
We employ a collection of metrics in the quantitative evaluation. Accuracy is measured by Mean Position Error (M*PE) and Mean Rotation Error (M*RE), reported separately for the Pelvis (P) versus other body Joints (J) to highlight the strong influence of the pelvis. We additionally break down the metrics for joints in the upper-body (UJ) versus the lower-body (LJ). We compute contact labels by thresholding the predicted probability at $0.5$ and compare them against ground truth.

\begin{table*}[h]
\begin{minipage}{\textwidth}
\centering
\begin{tabular}{cccccccccc}
\midrule
Model & MPPE${}_{cm}$ & MPRE${}^\circ$ & MJPE${}_{cm}$ & MJRE${}^\circ$ & Cont. Acc.${}_{\%}\uparrow$ & MUJPE${}_{cm}$ & MUJRE${}^\circ$ & MLJPE${}_{cm}$ & MLJRE${}^\circ$ \\
\midrule
\textit{LoBSTr$^{\dag}$} & $-$ & $-$ & $9.05$ & $12.09$ & $72.90$ & $8.08$ & $12.05$ & $15.69$ & $12.16$ \\
\textit{LoBSTr\_IMU$^{\dag}$} & $-$ & $-$ & $9.47$ & $12.97$ & $\underline{74.24}$ & $7.92$ & $12.62$ & $17.56$ & $13.99$ \\
\textit{AvatarPoser$^{\dag}$} & $\underline{9.69}$ & $\underline{12.31}$ & $8.73$ & $10.70$ & $-$ & $5.91$ & $9.61$ & $13.67$ & $12.60$ \\
\textit{AvatarPoser\_IMU$^{\dag}$} & $10.40$ & $12.77$ & $\underline{8.46}$ & $\underline{10.62}$ & $-$ & $\underline{5.42}$ & $\underline{9.08}$ & $\underline{13.10}$ & $\underline{11.90}$ \\
\textit{AvatarPoser}${\times}5^{\dag}$ & $12.25$ & $15.97$ & $9.65$ & $12.04$ & $-$ & $6.82$ & $11.37$ & $14.58$ & $13.23$ \\
DivaTrack & $\mathbf{6.73}$ & $\mathbf{9.96}$ & $\mathbf{7.35}$ & $\mathbf{9.26}$ & $\mathbf{85.25}$ & $\mathbf{4.78}$ & $\mathbf{8.13}$ & $\mathbf{11.86}$ & $\mathbf{11.23}$\\
\cdashline{1-10}
DivaTrack$^{*}$ & $9.99$ & $"$ & $9.10$ & $"$ & $"$ & $6.87$ & $"$ & $13.00$ & $"$ \\
\midrule
\end{tabular}
\caption{Comparison of retrained SOTA models and DivaTrack tested on DT Test dataset. The ${\dag}$ denotes retrained models on DT Train dataset and \textit{\_IMU} denotes that IMU accelerations are provided as additional input. DivaTrack$^{*}$ denotes results using the skeleton predicted from our calibration model, which has a mean bone-length error of ${1.7}_{cm}$. The $\bold{best}$ and \underline{runner-up} entries are annotated respectively.}
\label{tab:quansota}
\end{minipage}
\vspace{0.2cm}

\begin{minipage}{\textwidth}
\centering
\begin{tabular}{ccccccccccccccc}
\midrule
Chest & Chest2 & Chest3 & Chest4 & Neck & Head & Collar & Shoulder & Elbow & Wrist & Hip & Knee & Ankle & Toe\\
\midrule
0.55 & 1.11 & 1.01 & 0.90 & 1.54 & 1.07 & 1.00 & 1.22 & 1.94 & 1.1 & 0.45 & 4.97 & 2.95 & 2.02 \\
\midrule
\end{tabular}
\caption{Average bone length errors (cm) of each joint from our skeleton calibration model.}
\label{tab:skelcal}
\end{minipage}
\vspace{0.2cm}

\begin{minipage}{\textwidth}
\centering
\begin{tabular}{ccccccccc}
\midrule
\multirow{2}{*}{} & \multirow{2}{*}{TotalCapture} & \hspace{0.1cm} & \multicolumn{6}{c}{HPS} \\
\cline{4-9}
& & & BIB\_EG\_Tour & MPI\_EG & Working\_Standing & UG\_Computers & Go\_Around & UG\_Long \\
\midrule
\textit{AvatarPoser} & $11.96$ & & $22.53$ & $16.54$ & $19.08$ & $23.24$ & $19.5$ & $16.65$ \\
\textit{AGRoL} & $\mathbf{6.89}$ & & $28.95$ & $19.41$ & $17.67$ & $20.90$ & $14.16$ & $12.81$ \\
\textit{EgoPoser} & $-$ & & $9.55$ & $11.05$ & $8.70$ & $10.34$ & $6.9$ & $8.95$ \\
DivaTrack & $8.73$ & & $\mathbf{6.51}$ & $\mathbf{9.06}$ & $\mathbf{7.36}$ & $\mathbf{8.84}$ & $\mathbf{6.82}$ & $\mathbf{7.34}$ \\
\midrule
\end{tabular}
\caption{Comparison of Mean Joint Position Errors (MJPE${}_{cm}$) of released SOTA models and pretrained DivaTrack tested on TotalCapture and HPS, two public datasets with ground truth IMU sensor measurements.}
\label{tab:quanpublic}
\end{minipage}
\end{table*}

\begin{figure*}[ht]
  \centering
  \includegraphics[width=\linewidth]{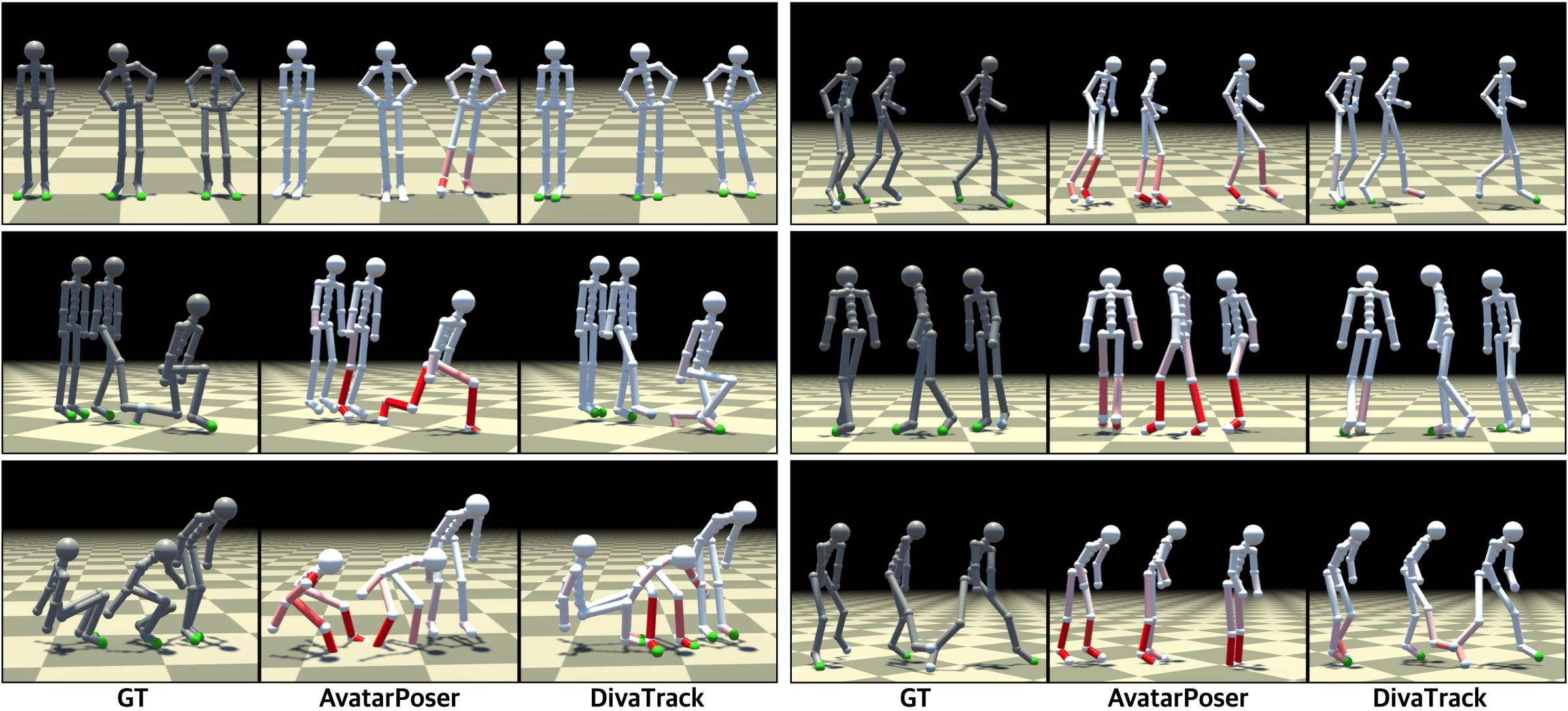}
  \caption{Qualitative comparison with AvatarPoser. The darker red color denotes a larger error. Our Model, DivaTrack, produces outputs closer to the ground truth for both static movements (left: hula-hoop, lunge, and sitting down) and various locomotion (right: running, moving backward, and moving a box) of subjects with different body shapes.}
  \label{fig:sotaqual}
\end{figure*}

\subsection{Comparison with State-of-the-art Methods}
We compare our method to state-of-the-art with a similar setup on both our own test set and public datasets with IMU sensors. We retrain baseline models on our training data for evaluation on our test set and use public pretrained models for evaluation on public datasets. 

On our own test set, we compare with recent methods that use sparse trackers on the upper body to generate full-body pose in real-time: \textit{LoBSTr}~\cite{yang21lobstr} and \textit{AvatarPoser}~\cite{jiang22avatarposer}. We also increase the \textit{AvatarPoser} model by 5x to be comparable with \textit{DivaTrack} model size as an additional baseline. Moreover, we augment both baseline models naively with linear accelerations from IMUs in the input with minimum change to their architectures, so they have access to the same information as \textit{DivaTrack}. All models are trained on our 18-subject training set, DT Train, and tested on our held-out dataset with 4 subjects, DT Test. All data used for training and test have preserved the skeleton offsets and joint rotations for each subject, without retargeting to a single standard proportion. As \textit{LoBSTr} requires an additional pelvis tracker, we provide it with ground truth pelvis transformation. We also provide the ground truth skeletons to both \textit{LoBSTr} and \textit{AvatarPoser} in all tests, and report results using ground truth and predicted skeletons respectively for \textit{DivaTrack}.

Quantitative results in Table \ref{tab:quansota} show that \textit{DivaTrack} with ground truth skeletons outperforms all baselines in all metrics. Our method notably improves over the runner-up entries in every metric, reducing MPPE by $2.96\text{cm}$ ($30\%$) and MJPE by \blue{$1.11\text{cm}$ ($13\%$)}, and reducing MPRE by $2.35^\circ$ ($19\%$) and MJRE by \blue{$1.36^\circ$ ($13\%$)}. In comparison, the bigger \textit{AvatarPoser}${\times}5^{\dag}$ model underperforms the original model, suggesting their architecture is not bottlenecked by network capacity. Augmenting existing models with IMU signals naively does not improve performance in a significant and conclusive way, hinting at the importance of our architectural decisions.

\textit{DivaTrack*} uses predicted skeletons and therefore its position accuracy degrades due to errors in bone offset estimation (Table \ref{tab:skelcal}). Our calibration model has an average error of $\mu=1.55_{cm},\ \sigma^2 = 1.18$ over all joints. Errors are smaller along the body trunk and larger in the limbs, where the knee has the largest error of $4.97cm$ since it is the most ambiguous based on the sensor signals. Despite these errors, \textit{DivaTrack*} still has better accuracy in the lower-body joint positions (MLJPE) than \textit{AvatarPoser$^{\dag}$} with ground truth skeletons.

We additionally evaluate our model in MPJPE on two public datasets with IMU sensors accelerations and ground truth body motion, TotalCapture \cite{trumble17totalcapture} (five subjects) and HPS \cite{guzov2021human} (seven subjects), as shown in Table \ref{tab:quanpublic}. We use ground truth skeletons in our method because we cannot run our calibration model on these datasets.

On TotalCapture, we compare \textit{DivaTrack} against published pretrained models of \textit{AvatarPoser} and \textit{AGRoL}. The dataset consists of walking, acting, and challenging freestyle activities, and we evaluate for the subset contained in AMASS \cite{mahmood19amass}. Even though \textit{AGRoL} demonstrates superior performance on average accuracy, its motions show severe jitter. Compared to \textit{AvatarPoser}, our model reduces MPJPE by $27\%$, consistent with results on our own test set.

On HPS, we compare \textit{DivaTrack} against \textit{AvatarPoser}, \textit{AGRoL}, and \textit{EgoPoser} \cite{jiang2023egoposer} with numbers reported in EgoPoser. The HPS dataset contains navigation and interaction movements in large-scale environments (ranging from $300 \sim 1000\textit{m}^2$, up to $2500\textit{m}^2$). Both \textit{AvatarPoser} and \textit{AGRoL} show significant performance decreases, because they use global representations of AMASS dataset for training, and cannot generalize well to large global offsets in outdoor environments. In contrast, \textit{DivaTrack} demonstrated robust and superior performance in large-scale scenes thanks to our reference frame representation.

Figure \ref{fig:sotaqual} are visual comparisons with \textit{AvatarPoser} on a few challenging examples for three-point input. For a variety of motion categories in both DT Test and TotalCapture, \textit{DivaTrack} follows the three-point signals closely with plausible leg movements and more clear foot contacts, while \textit{AvatarPoser} results are less accurate with more severe foot sliding artifacts. For more visual results, please refer to the supplementary video.

\subsection{Ablation of Reference Frame Definitions} 
\label{sssec:quanref} 

We compare the output of the P-Models using two respective reference frames, $R_{proj}$ and $R_{traj}$, and with the blended output from B-Model in Table \ref{tab:quanref} on DT Test. The blended result outperforms single reference results in position errors and contact accuracy. $R_{proj}$ performs better than $R_{traj}$ quantitatively, as the majority of test data has a strong correlation between head orientation and the movement direction. As shown in Figure \ref{fig:refqual} and in the supplementary video, $R_{traj}$ excels in cases when head movement doesn't correlate with body movement, such as occasionally turning to look back during backward walking. Our two-stream blending result outperforms each individual stream (P-Model), suggesting that it is able to adaptively combine the predictions to achieve the best result.

\begin{figure*}[ht]
  \centering
\includegraphics[width=\linewidth]{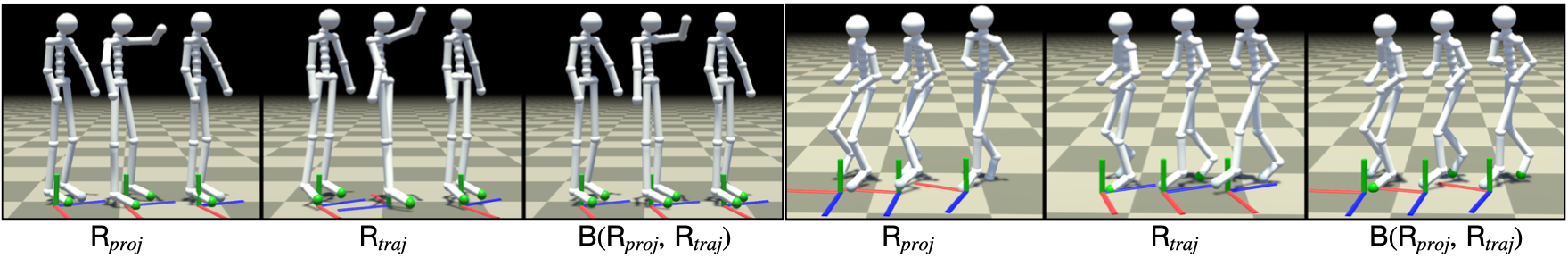}
  \caption{Comparison of reference frame definitions. Left: for static movements with slight head jiggling, $R_{traj}$ can rotate fast, resulting in jitter and sliding. Right: for locomotion with uncorrelated head direction, $R_{proj}$ are directed irrelevant to the translation, producing wrong poses. Our blending method can generate high-quality motions by taking advantage of two reference frames.}
  \label{fig:refqual}
\end{figure*}

\begin{figure*}[ht]
  \centering
\includegraphics[width=\linewidth]{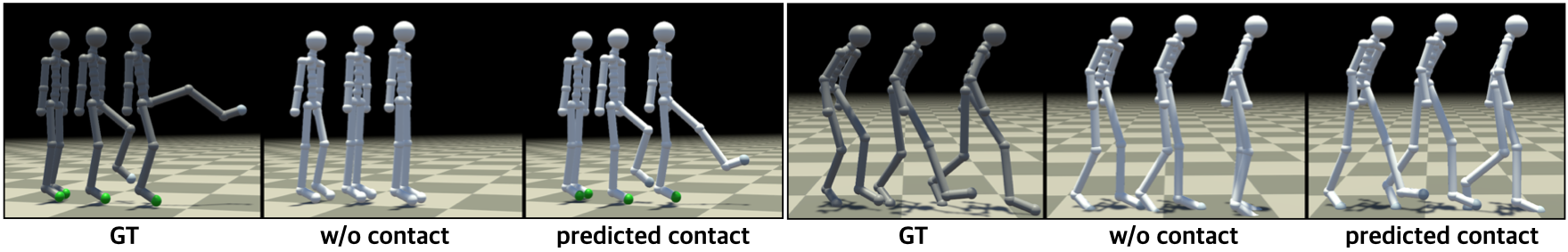}
  \caption{Comparison of contact labels as L-model condition. Foot contacts enable the L-model to generate motions that are underdetermined by three-point signals (i.e. kicking) and to maintain accurate footsteps while changing speed and direction.}
  \label{fig:contqual}
\end{figure*}

\begin{figure*}[ht]
  \centering
\includegraphics[width=\linewidth]{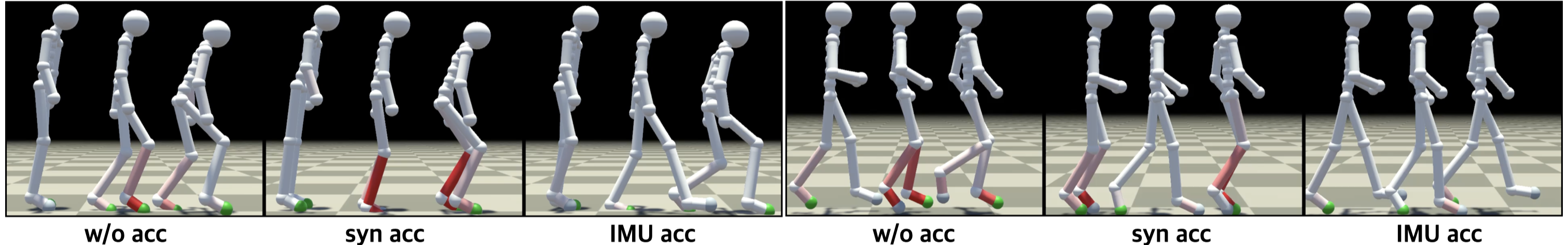}
  \caption{Comparison of acceleration sources. IMU accelerations help UC-Model to predict accurate contacts for corner cases (left: lunge and right: carrying object), thereby allowing L-Model to generate high-quality lower-body motions with minimized sliding artifacts.}
  \label{fig:imuqual}
\end{figure*}

\subsection{Ablation for Contacts as L-model Condition.} \label{sssec:quancont} 
We validate the utility of contact labels as a condition for the lower-body pose generation in ablation on DT Test. We trained two versions of the L-Model, one with contacts as L-Model condition and one without. We also compare using the predicted contact labels from the UC-model versus using the ground truth contact labels. The results in Table \ref{tab:quancont} clearly demonstrate that incorporating predicted contact labels as a condition improves the quality of generated lower-body poses and the quality has a clear correlation with the accuracy of predicted contact.

\begin{table}
\centering
\begin{tabular}{cccccc}
\midrule
Reference Def. & MPPE${}_{cm}$ & MJPE${}_{cm}$ & Cont. Acc.${}_{\%}\uparrow$ \\
\midrule
$R_{proj}$ & $7.19$ & $7.83$ & $81.44$\\
$R_{traj}$ & $7.51$ & $9.05$ & $82.34$\\
$R_{proj,\ traj}$ & $\mathbf{6.73}$ & $\mathbf{7.35}$ & $\mathbf{85.25}$\\
\midrule
\end{tabular}
\caption{Comparison of per-frame position errors for models trained with different reference definitions. $R_{proj,\ traj}$ denotes our two-stream blending approach of both reference frames.}
\label{tab:quanref}
\end{table}

\begin{table}
\centering
\begin{tabular}{ccc}
\midrule
 & MLJPE${}_{cm}$ & MLJRE${}^\circ$ \\
\midrule
w/o contacts & $13.27$ & $11.48$ \\
pred. contacts & $11.86$ & $11.23$ \\
GT contacts &  $\textbf{11.23}$ & $\textbf{10.63}$ \\
\midrule
\end{tabular}
\caption{Ablation on contact labels as L-Model condition (Mean Lower-body Joint Errors).}
\label{tab:quancont}
\end{table}

The improvement from contact prediction can be easily observed in the output motions as well. Figure \ref{fig:contqual} shows an example of kicking motion on the left, where foot contacts make it possible to distinguish standing from kicking with near identical upper-body postures. On the right is another example where contact information reduces foot sliding artifacts when walking speed changes.

\subsection{Ablation for IMU Acceleration Signals.} \label{sssec:quanimu} 
Given the importance of foot contact prediction, we incorporate linear acceleration signals from IMU sensors as an input to the UC-Model, because they can capture instantaneous contact events at high frequency, especially from the head. Compared with real IMU accelerations captured at up to 1000Hz, synthetic acceleration generated using finite differences is much lower frequency and less useful. We conducted an ablation to compare training the UC-Model without acceleration, with synthetic acceleration, and with IMU acceleration. We use two evaluation datasets: the full DT Test dataset, and a subset featuring challenging actions including Lunge, Kick, Lift knee, Turn/Walk/Run in place, and difficult Locomotion. Table \ref{tab:quanimu} clearly shows that IMU accelerations are essential for contact prediction and thus increase the performance of lower-body generation, while synthetic accelerations are not helpful as they do not provide new information.

While the improvements from IMU accelerations are more moderate in the full DT Test dataset because the majority of typical walking motions can already be well predicted without them, results from the challenging subset highlight where they shine. Figure \ref{fig:imuqual} showcases two examples where IMU accelerations provide a meaningful boost to pose accuracy. In the case of the lunge motion, only the model trained with IMU accelerations accurately captured the correct contact labels of the right foot leaving and landing on the ground, resulting in a realistic lunge of the lower body. Similarly, for the object-carrying motion, where the upper-body joints remain relatively stationary when compared to a standard walking motion, the utilization of IMU accelerations generated a more detailed walking motion of the lower body, while the other models exhibited static or incorrect poses. We found these classes of improvements make a significant difference in the overall visual quality of the method.

\begin{table}
\centering
\begin{tabular}{cccc}
\midrule
& $^{UC}$Cont. Acc.$_{\%}\uparrow$ & MLJPE${}_{cm}$ & MLJRE${}^\circ$ \\
\midrule
w/o acc & $78.47$ / $73.21$ & $12.03$ / $12.30$ & $11.53$ / $12.50$ \\ 
syn acc & $78.90$ / $73.95$ & $12.17$ / $12.41$ & $11.60$ / $12.60$ \\
IMU acc & $\mathbf{82.34}$ / $\mathbf{77.06}$ & $\mathbf{11.86}$ / $\mathbf{11.79}$& $\mathbf{11.23}$ / $\mathbf{11.90}$ \\
\midrule
\end{tabular}
\caption{Ablation on IMU linear accelerations as UC-Model input. The L-Model uses predicted contact labels from the UC-model as a condition for lower-body pose generation. The values on the left correspond to measurements taken across the entire DT Test set. The values on the right pertain specifically to a subset containing more challenging actions for the three-point tracking problem.}
\label{tab:quanimu}
\end{table}

\subsection{Real-time Tracking from VR devices and XSens IMUs}
We test \textit{DivaTrack} in real-time using a Meta VR device and three XSens IMU sensors (Figure \ref{fig:teaser}). One IMU is attached to the VR headset, and two IMUs are on the subject's wrists. After IMU calibration, we manually set the VR world to align with the IMU world. For real-time tracking, we fetch three-point positions and rotations from the VR device (Steam VR) and linear accelerations from IMU sensors (Xsens MVN). This data is streamed to Unity3D, where frame inputs are recorded and then sent to a Python inference module. \textit{DivaTrack} takes $7.5$ms per frame on a desktop with an RTX2080Ti GPU on average, which runs comfortably with $60$Hz VR data, although there is a noticeable system latency in our PCVR setup. We also run an online post-processing to smooth the output and stabilize foot contacts using IK. We tested subjects of various heights (in cm: 159, 162, 170, 172, 175, and 178) who are not in the training set, and instructed them to perform free-form locomotion. After the calibration poses, test subjects walked, jogged, and hopped back and forth, sat down on a stool or on the floor, and carried a stool to different locations (Figure~\ref{fig:teaser}). The output upper-body motions follow the three-point input signals well throughout all sessions, and the lower-body movements look plausible and coordinated with the upper-body, even though they are not always accurate. We invite readers to watch the supplementary video for qualitative evaluation.
\section{Discussion}
We demonstrated the pivotal role of reference coordinates in the final result. This is a unique challenge for three-point tracking problems due to the lack of root information. Our two-stream reference blending approach is effective but not elegant. Further, we still cannot effectively handle motions where the head's forward direction is close to the direction of gravity, such as lying down or crawling on the floor. A sub-optimal reference frame can lead to poor pelvis predictions that consequently cause severe body jitter and foot sliding artifacts. An unstable reference frame also makes it difficult to produce a static stance when the upper-body is moving. A deeper investigation of motion representation is needed for the three-point tracking problem.

We also demonstrated the importance of realistic IMU signals as opposed to synthetic accelerations for accurate contact prediction. TIP \cite{jiang2022tip} also observed similar discrepancies in these two types of signals. On the other hand, the three-point data can be realistically simulated from motion capture data, as shown in QuestSim \cite{winkler2022questsim}. As a result, we collected our dataset with XSens similar to TotalCapture but did not evaluate our method on public datasets with no IMU signals. If VR devices can expose their IMU data, a realistic dataset of three-point data and IMU signals will be tremendously helpful to future research.

Our method encounters challenges with unseen motion categories, notably highlighted in our supplementary video. In addition, as our training data is confined to the flat ground (height=0) that defines reference coordinates, our model struggles with motions involving ascent and descent.

While our synthetic tracker data aligns well with real MR device tracking signals, it fails to capture tracking failures in practical VR/MR usage, and our method overlooks this issue. Addressing the challenge of ``tracking loss'' in practical scenarios will be an intriguing avenue for future research.

Our framework currently predicts the user skeleton and full-body pose separately, before combining them at the last stage. A more effective design would be to integrate them earlier so the pose generation model can utilize the skeleton proportion in pose prediction. Unfortunately, our preliminary test did not yield any significant improvements.
\section{Conclusion}
We presented DivaTrack, a deep learning framework that infers a user's motion in real-time using six DOF poses and linear accelerations from three trackers on the head and wrists. Considering the different level of information from the sensors, DivaTrack first infers upper-body poses and contact states, and use them as control signals to generate suitable lower-body poses. Addressing the lack of known pelvis joint information, our model generates poses with respect to two different head-based reference frames and blends them to create a final pose. From a simple set of calibration poses, DivaTrack infers a user-specific skeleton model for final visualization. Our results suggest that signals from the head and hands alone can provide rich information about body proportions and motion. Our insights about the importance of IMU acceleration signals and their correlation to foot contacts are validated by ablation studies. We further provide a dedicated dataset to showcase challenges of using only three-point inputs, and we believe it can inform future research to tackle fundamental issues in this problem.
\printbibliography
\end{document}